\documentclass[10pt,twocolumn,letterpaper]{article}

\usepackage{3dv}
\usepackage{times}
\usepackage{epsfig}
\usepackage{graphicx}
\usepackage{amsmath}
\usepackage{amssymb}

\usepackage{graphics} 
\usepackage{epsfig} 
\usepackage{mathptmx} 
\usepackage{color}
\usepackage{subcaption}
\usepackage{tabularx}
\usepackage{makecell}
\usepackage{multicol}
\usepackage{mwe,lipsum}
\usepackage{array,multirow}
\usepackage{float}
\usepackage{eucal}

\newcommand{\rr}[1]{{\textcolor{red}{\textbf{#1}}}}

\newcommand{\bb}[1]{{\textcolor{cyan}{\textbf{#1}}}}
\newcommand{\bl}[1]{{\textcolor{black}{\textbf{#1}}}}

\usepackage[pagebackref=true,breaklinks=true,colorlinks,bookmarks=false]{hyperref}

\threedvfinalcopy 

\def\threedvPaperID{308} 

\ifthreedvfinal\fi
\begin{document}

\title{Modality-Guided Subnetwork for Salient Object Detection}

\author{Zongwei Wu$^{1,3}$ \quad Guillaume Allibert $^{2}$ \quad Christophe Stolz$^{1}$ \quad Chao Ma$^{3}$ \quad C\'edric Demonceaux $^{4}$ \thanks{~Corresponding author.} \\

${}^{1}$ VIBOT EMR CNRS 6000, ImViA, Universit\'e Bourgogne Franche-Comt\'e, France\\
${}^{2}$ Universit\'e  C\^ote d’Azur, CNRS, I3S, France\\
${}^{3}$ MoE Key Lab of Artificial Intelligence, AI Institute, Shanghai Jiao Tong University, China \\
${}^{4}$ ImViA, Universit\'e Bourgogne Franche-Comt\'e, France\\

{\tt\small zongwei\_wu@etu.u-bourgogne.fr, allibert@i3s.unice.fr, chaoma@sjtu.edu.cn} \\ {\tt\small\{christophe.stolz,cedric.demonceaux\}@u-bourgogne.fr }}

\maketitle
\thispagestyle{empty}


\begin{abstract}
   Recent RGBD-based models for saliency detection have attracted research attention. The depth clues such as boundary clues, surface normal, shape attribute, etc., contribute to the identification of salient objects with complicated scenarios. However, most RGBD networks require multi-modalities from the input side and feed them separately through a two-stream design, which inevitably results in extra costs on depth sensors and computation. To tackle these inconveniences, we present in this paper a novel fusion design named modality-guided subnetwork (MGSnet). It has the following superior designs: 1) Our model works for both RGB and RGBD data, and dynamically estimates depth if not available. Taking the inner workings of depth-prediction networks into account, we propose to estimate the pseudo-geometry maps from RGB input — essentially mimicking the multi-modality input. 2) Our MGSnet for RGB SOD results in real-time inference but achieves state-of-the-art performance compared to other RGB models. 3) The flexible and lightweight design of MGS facilitates the integration into RGBD two-streaming models. The introduced fusion design enables a cross-modality interaction to enable further progress but with a minimal cost.
\end{abstract}

\section{Introduction}

\begin{figure}[ht]
\centering
\includegraphics[width=\linewidth,keepaspectratio]{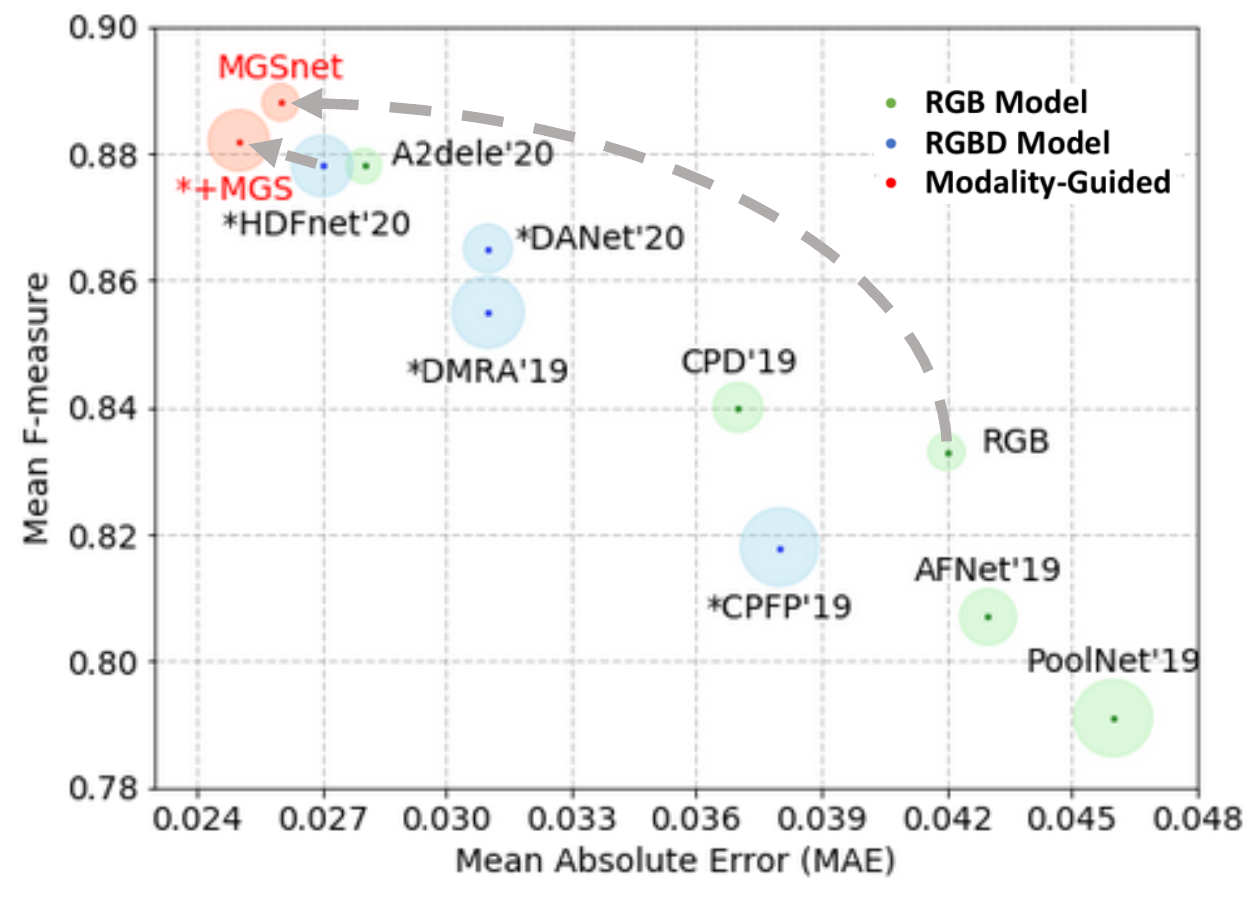}
\vspace{-7mm}
\caption{Performance analysis on NLPR dataset \cite{peng2014rgbd}. Note that better models are shown in the upper left corner (i.e., with a larger mean F-measure and smaller MAE). The circle size denotes the model size. Our proposed MGSnet for RGB SOD achieves the best performance with the lightest model size. The MGS design can also be embedded to the state-of-the-art RGBD model HDFnet \cite{pang2020hierarchical} to enable further progress (denoted as $*+MGS$).}
\label{fig:perform}
\vspace{-5mm}
\end{figure}

\begin{figure*}[ht]
\centering
\includegraphics[width=\linewidth,keepaspectratio]{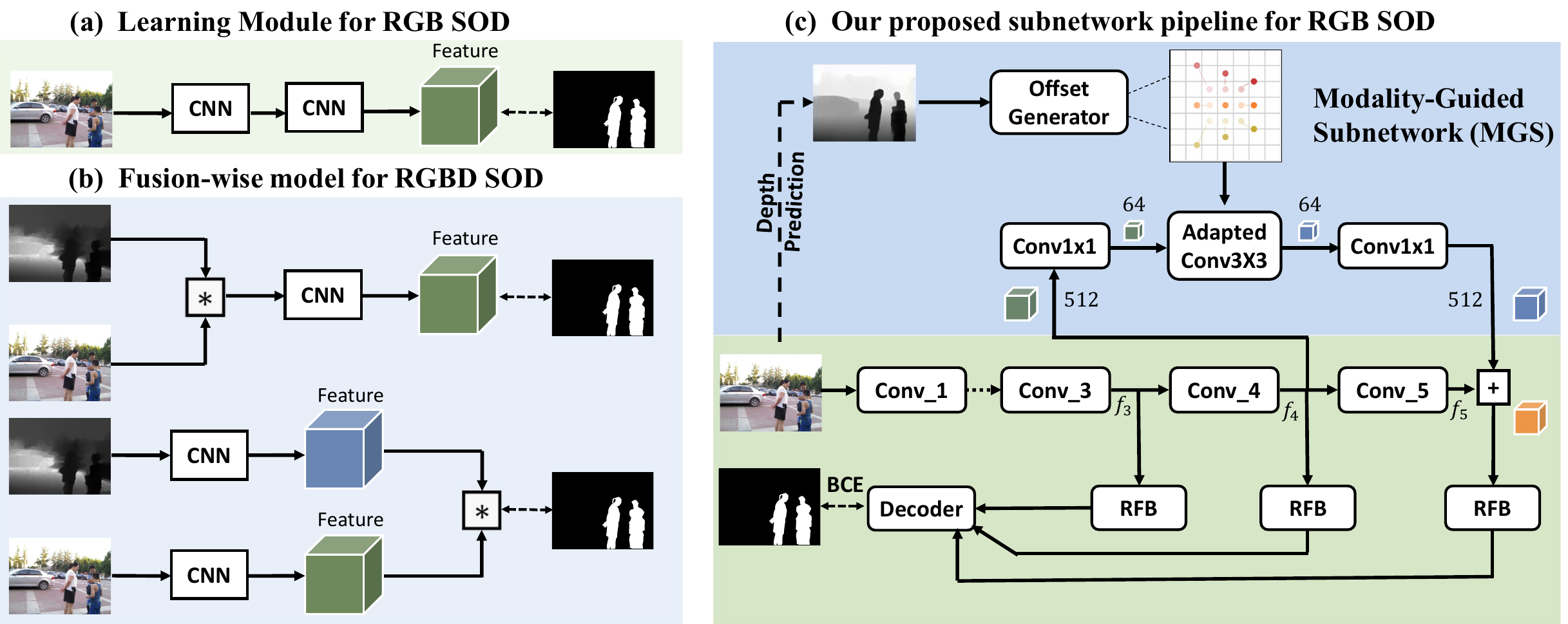}
\vspace{-5mm}
\caption{(a) Conventional RGB models \cite{liu2019simple,zhang2017learningsingle,wu2019cascaded} insert additional modules to learn geometry-invariant features. (b) RGBD models \cite{fu2020jldcf,zhao2020single,pang2020hierarchical} adopt fusion-wise design to learn both photometric and geometric information. (c) Our proposed MGSnet which takes only RGB image for both training and testing. We use depth prior to guide sampling position on RGB feature map through a subnetwork design to compensate the master streaming.}
\label{fig:MGS}
\vspace{-5mm}
\end{figure*}

In the last decade, RGB-based deep learning models for salient object detection (SOD) \cite{zhang2017learningsingle,deng2018r3net,liu2019simple,wu2019cascaded,zhao2019egnet} achieved significant success thanks to the advances of GPU and CNN. Given an input image, the goal of SOD is to compute the pixel-wise location of the prominent objects that visually attract human attention the most. However, RGB SOD models focus more on photometric information instead of geometry. This is due to the fixed shape and size kernel design of CNN that is not invariant to scale changes and to 3D rotations. By the lack of geometric information on the input side, it is inevitable for RGB models to add additional learning modules in the network to attend to salient objects, resulting in model complexity and computational cost.

Recent RGBD-based SOD has motivated research interest thanks to the accessibility of cross-modal information from the input side. State-of-the-art RGBD models \cite{fu2020jldcf,piao2020a2dele,pang2020hierarchical,zhao2020single} achieve superior performance over the RGB baseline, affirming the effectiveness of learning from two modalities. Most architectures adapt fusion-wise models, such as early fusion \cite{zhao2020single} where the depth map is fed as the fourth channel to RGB image, or multi-scale and late fusion \cite{pang2020hierarchical} where two-stream networks are adopted. However, early fusion contains more low-level features than semantic ones. Multi-scale or late fusion inevitably requires more learning parameters. As shown in Figure \ref{fig:perform}, the size of RGBD models is often larger than that of RGB networks. 


We explore differently the relationship between depth map and RGB image. Taking human beings as an example, to distinguish salient objects from the 3D world, the input is the visual appearance through human eyes. With the color information and thanks to the depth estimation capability, humans further discover geometric information. This prior guides the understanding of RGB images. It should be the same case for intelligent machines. 

To this end, we propose a novel Modality-Guided Subnetwork (MGSnet) which adaptively transforms convolutions by fusing information from one modality to another (e.g., depth to RGB or RGB to depth). Our network matches perfectly both RGB and RGB-D data and dynamically estimates depth if not available by simply applying an off-the-shelf depth prediction model. We design a subnetwork mechanism alongside the master streaming pipeline. The subnetwork can be treated like a light residual-addition branch as the ResNet \cite{He2016Residual}. It takes one modality map as the master input, e.g. RGB, and enhances its robustness by deforming the convolution kernel with the supervision of the complementary modal prior, e.g. depth, and vice versa.

In summary, the main contributions of this paper are listed as follows : 
\begin{itemize}
\vspace{-2mm}
\setlength{\itemsep}{2pt}
\setlength{\parsep}{0pt}
\setlength{\parskip}{0pt}
    \item  By exploiting the nature of CNN sampling position, we propose a novel cross-modal fusion design (MGS) for salient object detection, where we use a subsidiary modality, i.e., RGB/depth, to guide the main modality streaming, i.e., depth/RGB. 
    \item For RGB-only input, we suggest using an off-the-shelf depth prediction model to mimick the multi-modality input. Our MGSnet enables dramatical performance gain on benchmark datasets and achieves state-of-the-art performance among RGB SOD models.
    \item The proposed MGS can also be embedded in RGBD two-stream network with the advantage of cross-modality cues while being lightweight.
\end{itemize}

\section{Related Work}
\textbf{RGB SOD:} In the past decade, the development of GPU and CNN contributes to the advances of RGB SOD. One core problem is understanding the geometric information from the image. Fully Convolutional Network (FCN) \cite{Long2015FCN} is a pioneering work in leveraging spatial information in CNN. Most recent researches dominating RGB SOD are FCN-based, such as \cite{zhang2017learningsingle} which designs a single stream encoder-decoder system, \cite{li2016deepmulti} which adopts a multi-scale network on input, and most currently \cite{deng2018r3net,liu2019simple,wu2019cascaded,zhao2019egnet} which fuse multi-level feature map. Some branch designs also have achieved impressive results such as C2S-Net \cite{li2018contour} which bridges contour knowledge for SOD. By inserting additional transformation parameters in networks, it contributes to the model performance. Nevertheless, the inference time and computational cost become more significant.


\textbf{RGBD SOD:} The complementary depth map may provide extra clues on the geometry. How to efficiently joint RGB and depth modality is the key challenge for RGBD SOD. One possible solution is to treat the depth map as an additional channel and adapt a single-stream system as shown in DANet \cite{zhao2020single}. It further designs a verification process with a depth-enhanced dual attention module. An alternative is to realize multi-stream networks followed by a feature fusion mechanism. PDNet \cite{zhu2019pdnet} designs a depth-enhanced stream to extract geometric features and further fuses with the RGB features. D3net \cite{fan2020rethinkingd3} adopts separate networks to respectively extract features from RGB, depth map, and RGBD four-channel input. A late fusion is further realized. HDFnet \cite{pang2020hierarchical} adopts two streaming networks for both RGB image and depth map. These features are further fused to generate region-aware dynamic filters. JL-DCF \cite{fu2020jldcf} proposes joint learning from cross-modal information through a Siamese network. Generally, RGBD networks achieve superior performance compared to RGB as shown in Figure \ref{fig:perform}. However, these methods rely on the quality and accessibility of the depth map. A high-quality depth map requires expensive depth sensors and is still sparse compared to an RGB image as suggested in \cite{fu2020jldcf,fan2020rethinkingd3}. To this end, DCF \cite{ji2021calibrated} proposes to calibrate the raw depth to improve the quality. Nevertheless, the high computational cost due to the two-streaming network requires more development.

Some recent researches \cite{piao2020a2dele,ji2020accurate,zhao2020depth} propose to learn from RGBD images and tests on RGB. This design enables an RGB CNN to achieve a comparable result with RGBD SOD during testing. Different from it, we propose to firstly discover the hidden geometric modality behind RGB images by simply using an off-the-shelf depth prediction method. With the estimated depth, we further propose a Modality-Guided Subnetwork mechanism to enhance the master RGB network understanding of the contour problem. Our proposed MGSnet achieves state-of-the-art performance with real-time inference speed compared to other RGB models. It can also be embedded in RGBD two-stream models to enable further progress with raw depth.

\section{Modality-Guided Subnetwork}
\subsection{Overview}

In Figure \ref{fig:MGS} (c), our network only takes RGB as input that then estimates the pseudo-depth. Our MGSnet only takes the pseudo-depth to deform the RGB streaming. In other words, only the RGB modality is fed through Conv\_4. 

\begin{figure}[ht]
\centering
\includegraphics[width=\linewidth,keepaspectratio]{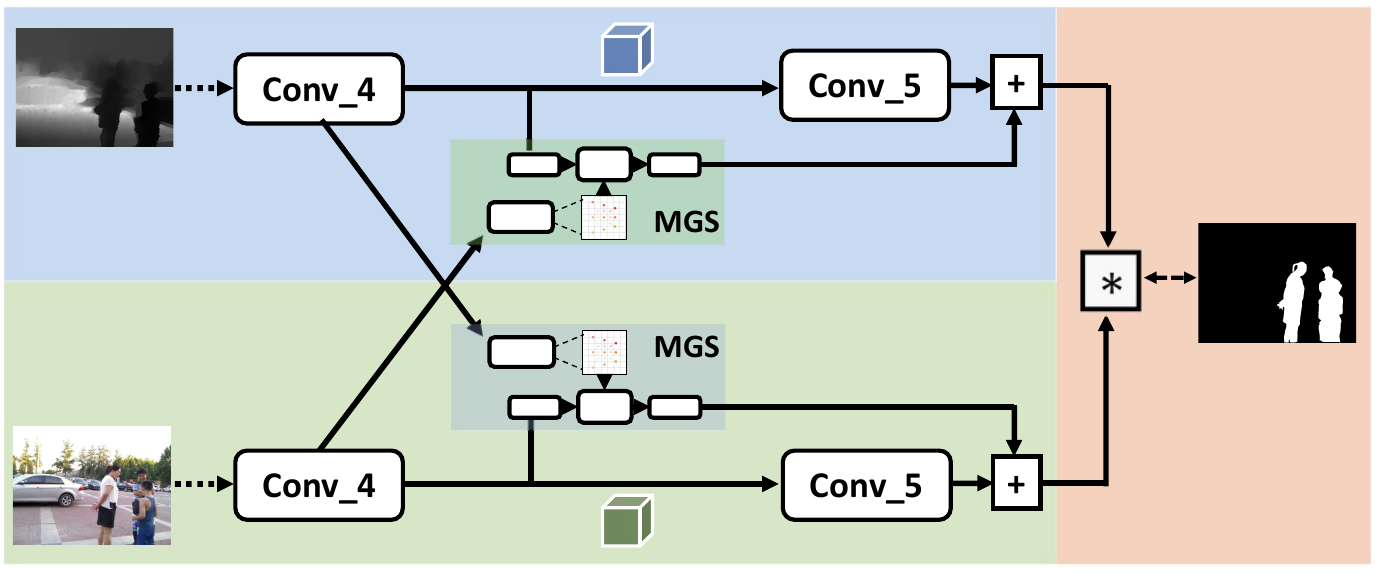}
\vspace{-7mm}
\caption{Illustration of embedded MGS on a RGBD two-streaming network.}
\label{fig:rgbdmgs}
\vspace{-5mm}
\end{figure}

\begin{figure*}[ht]
\begin{subfigure}{.495\textwidth}
\centering
\includegraphics[width=0.98\linewidth]{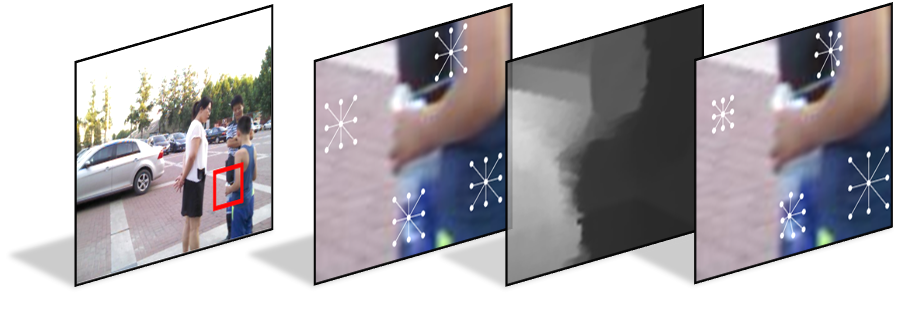}
\caption{Depth-guided sampling position on RGB image}
\end{subfigure}
\begin{subfigure}{.495\textwidth}
\centering
\includegraphics[width=0.98\linewidth]{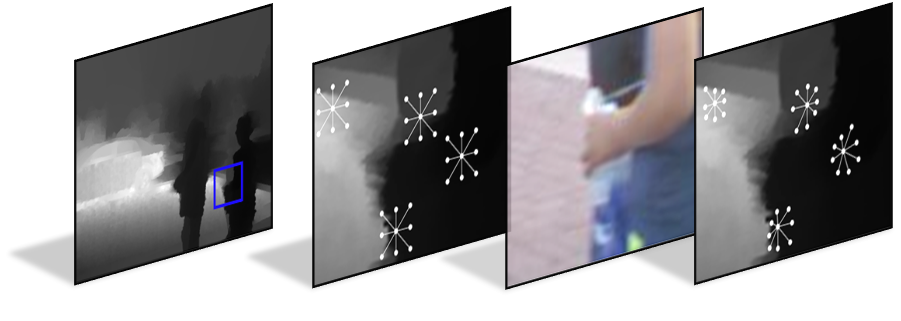}
\caption{RGB-guided sampling position on depth image}
\end{subfigure}
\vspace{-3mm}
\caption{Visual comparison of MGSnet. A pair of RGB and depth images from an RGBD dataset are illustrated on the left. While extracting features through two streaming networks, the cross-modal information beyond the fixed sampling position is not utilized (second left). Our proposed modality-guided sampling position breaks the limit of fixed-local configurations. The new sampling position incorporates supporting modality into the basic function of CNN on the main modality: the fixed sampling position is replaced by relevant neighbors defined by the supporting modality without limitation (right).}
\label{fig:guided}
\vspace{-3mm}
\end{figure*}

Note that our model is not limited by the nature of the modality. It can be a depth-guided RGB convolution as well as an RGB-guided depth convolution. Figure \ref{fig:rgbdmgs} presents our model embedded on an RGBD two-streaming network and Figure \ref{fig:guided} illustrates the idea of modality-guided sampling position. We learn the offset from both semantic RGB and depth features to create a cross supervision mechanism. 

For simplicity, we present in the following section a depth-guided subnetwork for RGB features. It contains three parts:  a master RGB streaming network, an off-the-shelf prediction model to estimate a pseudo-depth map if not available, and a depth-guided subnetwork design. For simplicity, VGG-16 \cite{simonyan2014vgg} architecture is adopted as our basic convolutional network to extract RGB features for its wide application in SOD. We use RFB \cite{liu2018rfb} on the steamer layers ($f_3, f_4, f_5$) which contains high level features for SOD as suggested in \cite{piao2020a2dele,pang2020hierarchical,fu2020jldcf}. We further embed our subnetwork to enhance the edge understanding of the encoder output. We take the same decoder as proposed in \cite{piao2020a2dele} and a simple binary cross-entropy (BCE) as the loss. 


\subsection{Depth-guided Subnetwork}
To proceed with the geometric prior, the depth map D and the RGB feature map (output of $Conv\_4$) are fed together to our model. We use $f_4 \in \mathbb{R}^{b\times 512\times h \times w}$ to denote the input RGB feature. The depth prior and RGB feature maps are articulated through an adaptive convolution to compute depth-aware RGB feature maps as output. The last is added to the master RGB stream to form the final feature map.

The subnetwork contains three convolutions of different filter sizes: $1\times1, 3\times3$, and $1\times1$. It shares the same architecture of plain baseline of ResNet \cite{He2016Residual} that the $1\times1$ layers are used for reducing (512$\rightarrow$64) and then increasing dimensions (64$\rightarrow$512), allowing the $3\times3$ layer with smaller input/output dimensions. We denote $\mathcal{D}$ and $\mathcal{U}$ for the first and the last $1\times1$ convolution, which stands for down-sample and up-sample, respectively. This design can significantly reduce the learning parameters, which contributes to the lightweight design of our subnetwork. Different from ResNet that uses the three layers as a bottleneck, we use them as the residual-addition branch which serves as complementary information to the plain network. 

A standard convolution is formulated as: 
\begin{equation}
   \textbf{y}(p) = \sum_{p_n\in\textbf{R(p)}} \textbf{w}(p_n)  \cdot \textbf{x}(p + p_n).
\end{equation}
with \textbf{R(p)} the sampling grid for point $p$. Physically it represents a local neighborhood on input feature map, which conventionally has regular shape with certain dilation $\Delta d$, such that :
\begin{equation}
    \textbf{R(p)} = a \Vec{u} + b\Vec{v}
    \label{eq:2dgrid}
\end{equation}
where $(\Vec{u}, \Vec{v})$ is the pixel coordinate system of input feature map and $(a,b) \in (\Delta d \cdot \{-1, 0 , 1\})^2$ for a $3\times3$ convolution or $(a,b) \in (\{0\})^2$ for $1\times1$ convolution. 

We further replace the conventional $3\times3$ convolution by deformable convolution (DeformConv) \cite{dai2017deformable}, where the kernels are generated with different sampling distributions which is adapted to depth modality. Mathematically, we have:

\begin{equation}
   \textbf{y}(p) = \sum_{p_n\in\textbf{R(p)}} \textbf{w}(p_n)  \cdot \textbf{x}(p + p_n + \Delta p_n)
   \label{eq}
\end{equation}

The convolution may be operated on the irregular positions $p_n+\Delta p_n$ as the offset $\Delta p_n$ may be fractional. To address the issue, we use the bilinear interpolation which is the same as that proposed in \cite{dai2017deformable}. The adapted convolution is denoted as $\mathcal{A}$.

Thanks to the depth input of the subnetwork, the scale and geometric transformation of objects on the RGB feature map can be directly analyzed with the adapted offsets. This process is expressed as:

\begin{equation}
   \Delta p_n = \eta (D)
   \label{eq:offset}
\end{equation}

We present two types of offset generators according to different plain networks. More details are discussed in the following section. The newly defined sampling position becomes depth-aware and helps to better articulate the RGB feature and geometric information. Finally, the output of MGS is added to the master stream, which serves as complementary depth-aware guidance on RGB features.

The entire process to compute the modality-guided feature $f_M$ can be formulated as follows:
\begin{equation}
\begin{split}
f_M & = MGS(f_4, D) \\
    & = \mathcal{U}(\mathcal{A}(\mathcal{D}(f_4), \eta(D)))\\
\end{split}
\end{equation}

The output of RGB encoder can be formulated as :
\begin{equation}
   out = f_5 + \lambda f_M
  \label{eq:reg}
\end{equation}
where $\lambda$ is the weight parameter.

\subsection{Offset generator}

We use another modal prior to deform the main stream convolution. When the offset exceeds the input size, the output will be computed as if the zeros padding is applied. For RGB input, the pseudo-depth is used to deform the RGB sampling position. The offset is generated through Z-ACN \cite{wu:zacn}. It firstly back-projects the 2D conventional grid to form a 3D point cloud according to the depth. Based on the point cloud, it extracts a depth-aware 3D plan and further creates a depth-aware 3D regular grid. Then it projects the 3D regular grid to the image plan to form the deformable sampling position. More details can be found in Z-ACN \cite{wu:zacn} paper. Different to DeformConv \cite{dai2017deformable} that learns offset from the RGB feature map to deform RGB sampling position, Z-ACN computes offset according to low-level geometric constraint (one-channel depth) and does not require gradient descent, thus perfectly matches our light-weight subnetwork design. The computed offset allows the RGB convolution to be scale and rotation independent. We verify through experiments the superior performance of our model in the ablation study.

For RGBD input, current Sconv \cite{chen2021spatial} suggests learning the RGB offset from a semantic depth feature map. We share the same motivation as Sconv. However, Sconv firstly projects the depth into a high-dimensional feature space and secondly learns a depth-aware offset and mask. Unlike Sconv, we learn the offset from the encoder or high-level features to avoid the additional projection. In other words, in our case, the offset generator $\eta$ is realized through a simple $3\times3$ convolution to minimize the computational cost. Furthermore, we adapt to different modalities as input, i.e., it learns offset from both RGB and depth, while Sconv only learns from depth.  

\subsection{Understand adaptive sampling position}

\begin{figure*}[ht]
\centering
\includegraphics[width=\linewidth,keepaspectratio]{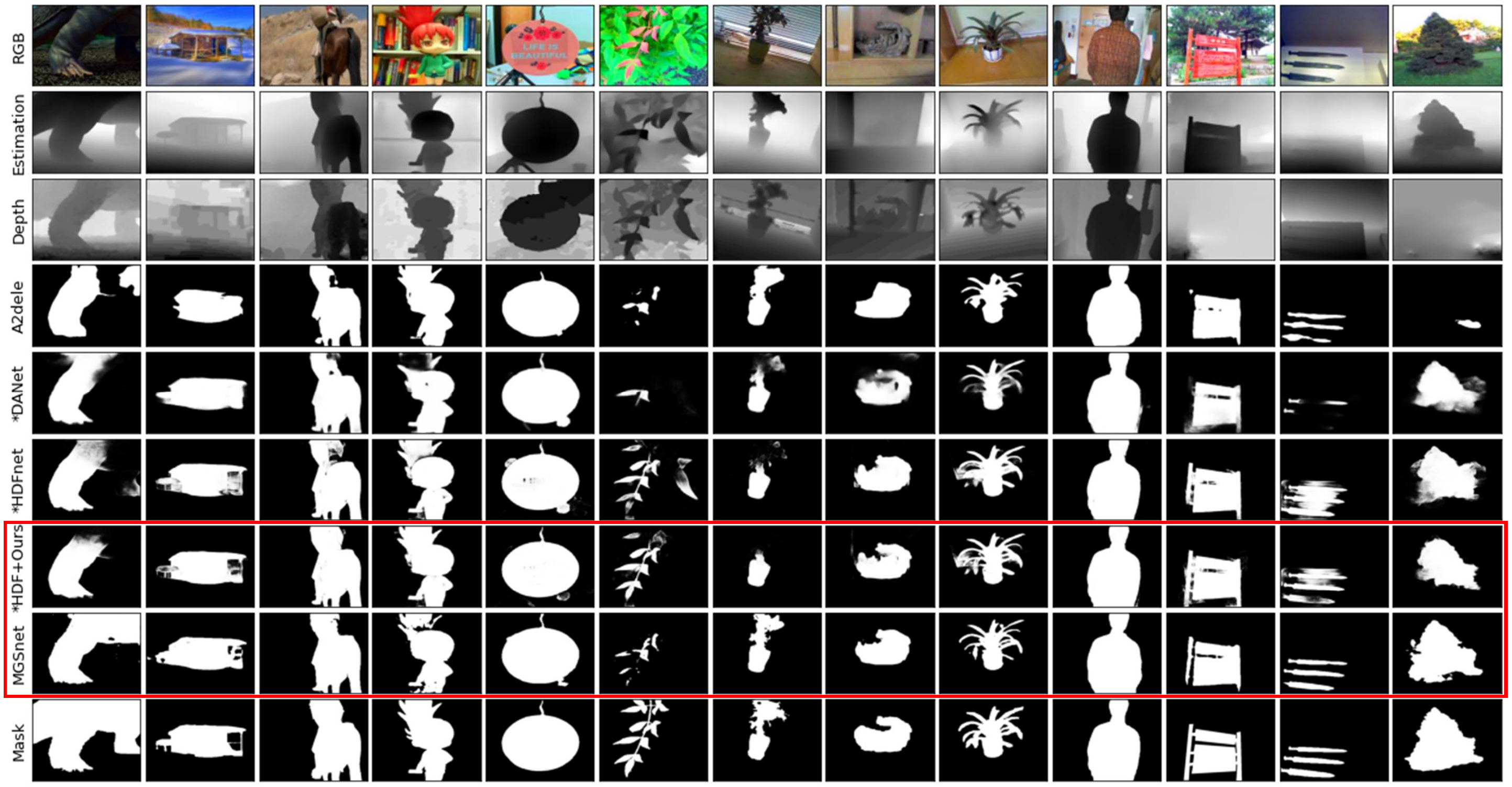}
\vspace{-6mm}
\caption{Visual comparison between our proposed MGSnet and the state-of-the-art RGB/RGBD methods.  $*$ denotes that the ground truth depth is used during testing. We also embed MGS on the HDFnet \cite{pang2020hierarchical} to enable further improvement, denoted as $*HDF+Ours$. More examples can be found in the Supplementary Material.} 
\label{fig:rgb}
\vspace{-8mm}
\end{figure*}

Our model aims to compensate for the single modality streaming. As shown in Figure \ref{fig:guided}, while extracting features from RGB images, the conventional sampling position is limited by the lack of capability to include geometry due to the fixed shape. We propose to use the depth prior to accurately locate the sampling position. For RGB input without depth prior, we suggest mimicking the depth map by using a monocular depth estimation model. Some pseudo-depth images may be inaccurate due to the domain gap between SOD and monocular depth estimation. In such a case, the offset will converge to 0 so that the deformation becomes minimal and local. The contribution of the depth-aware RGB feature is further regularized by the weight parameter $\lambda$ of Eq. \ref{eq:reg}. In Fig. \ref{fig:rgb}, we show that our method is robust to non-optical depth through several examples.

While extracting features from raw depth, conventional sampling positions may produce sub-optimal results due to some inaccurate measurements. The raw depth maps for SOD are obtained by camera measurements such as Kinect and Light Field cameras, or estimated by classic computer vision algorithms as \cite{sun2010secrets,liu2010sift}. Thus, the raw depth images may contain noise and ambiguity. We can visualize several low-quality samples on the third row of Figure \ref{fig:rgb}. To this end, we propose to use the RGB image to deform the depth sampling position. In such a case, the RGB-guided sampling position can make up for the measurement error on geometry.

\section{Experiments}

\subsection{Benchmark Dataset}
To verify the effectiveness of our method, we conduct experiments on seven following benchmark RGBD datasets. DES \cite{cheng2014depth} : includes 135 images about indoor scenes captured by Kinect camera. LFSD \cite{li2014saliency}: contains 100 images collected on the light field with an embedded depth map and human-labeled ground truths. NLPR \cite{peng2014rgbd}: contains 1000 natural images captured by Kinect under different illumination conditions. NJUD \cite{ju2014depth}: contains 1,985 stereo image pairs from different sources such as the Internet, 3D movies, and photographs taken by a Fuji W3 stereo camera and with estimated depth by using optical flow method \cite{sun2010secrets}. SSD \cite{zhu2017ssd}: contains 80 images picked up from stereo movies with estimated depth from flow map \cite{sun2010secrets}. STEREO \cite{niu2012leveraging}: includes 1000 stereoscopic images downloaded from the Internet where the depth map is estimated by using SIFT flow method \cite{liu2010sift}. DUT-RGBD \cite{piao2019depth}: contains 1200 images captured by Lytro camera in real-life scenes.


\subsection{Experimental Settings}

Our model is implemented basing on the Pytorch toolbox and trained with a GTX 3090Ti GPU. We adopt several generally-recognized metrics for quantitative evaluation: F-measure is a region-based similarity metric that takes into account both Precision (Pre) and Recall (Rec). Mathematically, we have : $F_{\beta} = \frac{(1+\beta^2) \cdot Pre \cdot Rec}{ \beta^2 \cdot Pre + Rec}$. The value of $\beta^2$ is set to be $0.3$ as suggested in \cite{achanta2009frequency} to emphasize the precision. In this paper, we report the \textbf{maximum F-measure} ($F_{\beta}$) score across the binary maps of different thresholds, the \textbf{mean F-measure} ($F_{\beta}^{mean}$) score across an adaptive threshold and the \textbf{weighted F-measure} ($F_{\beta}^{w}$) which focuses more on the weighted precision and weighted recall.
\textbf{Mean Absolute Error ($MAE$)} studies the approximation degree between the saliency map and ground-truth map on the pixel level.
\textbf{S-measure ($S_m$)} evaluates the similarities between object-aware ($S_o$) and region-aware ($S_r$) structure between the saliency map and  ground-truth map. Mathematically, we have: $S_m= \alpha \cdot S_o + (1 - \alpha) \cdot S_r$, where $\alpha$ is set to be $0.5$.
\textbf{E-measure ($E_m$)} studies both image level statistics and local pixel matching information. Mathematically, we have: $E_m= \frac{1}{W\times H} \sum_{i=1}^{W} \sum_{j=1}^{H} \phi_{FM}(i,j)$, where $\phi_{FM}(i,j)$ stands for the enhanced-alignment matrix as presented in \cite{fan2018enhanced}.


\subsection{Performance Comparison with RGB Input}

\begin{table*}
\setlength\tabcolsep{0pt}
\setlength\extrarowheight{1pt}
\begin{center}
\begin{tabular*}{0.98\textwidth}{@{\extracolsep{\fill}}*{18}{c}}
\hline
 Dataset   & &\multicolumn{3}{c}{DES} & \multicolumn{3}{c}{NLPR}  & \multicolumn{3}{c}{NJUD}  & \multicolumn{3}{c}{STEREO} & \multicolumn{3}{c}{DUT-RGBD}  \\
\cline{3-5} \cline{6-8} \cline{9-11} \cline{12-14} \cline{15-17}
Metric & \small $Size\downarrow$ & \small $MAE\downarrow$ & \small $F_{\beta}^{mean}\uparrow $  & \small $F_{\beta}^{w}\uparrow$ & \small $MAE\downarrow$ &\small $F_{\beta}^{mean}\uparrow $  & \small $F_{\beta}^{w}\uparrow$ &\small $MAE\downarrow$ & \small $F_{\beta}^{mean}\uparrow $  & \small $F_{\beta}^{w}\uparrow$ & \small $MAE\downarrow$ & \small $F_{\beta}^{mean}\uparrow $  & \small $F_{\beta}^{w}\uparrow$ & \small $MAE\downarrow$ & \small $F_{\beta}^{mean}\uparrow $  & \small $F_{\beta}^{w}\uparrow$ \\
\hline
\multicolumn{17}{c}{RGB input}
\\
$R^{3} Net_{18}$ 
 &225  &.066 &.728 &.693   &.101 &.649 &.611   &.092 &.775 &.736  &.084 &.800 &.752    & .113 &.781 &.709\\
$PoolNet_{19}$  
 &279  &.031 &.852 &.814   &.046 &.791 &.771   &.057 &.850 &.816  &.045 &.877 &.849  & .049 &.871 &.836   \\
$CPD_{19}$      
 &112  & \rr{.028} &.860 &\rr{.841}     &.037 &.840 &.829     &.059 &.853 &.821  &.046 &.880 &.851  & .055 &.872 &.835 \\
$AFNet_{19}$    
 &144     & .034 &.840 &.816       &.043 &.807 &.796     &.056 &.857 &.832  &.046 &.876 &.850  & .064 &.851 &.817   \\
$EGNet_{19}$    
&412     & .035 &.831 &.797       &.047 &.800 &.774     &.060 &.846 &.808    &.049 &.876 &.835  & .059 &.866 &.805   \\
\multicolumn{17}{c}{+ Pseudo Depth (86 Mb extra model size)} \\

$HDFnet_{20}$ &177 (+86)     
& .070 &.721 &.664       &.062 &.758 &.741     &.124 &.716 &.656  & .106 &.743 &.684   & - & - & -  \\
$CoNet_{20}$ &171 (+86)     
& .037 &.820 &.808       &.049 &.744 &.835     &.068 &.827 &.795  & .050 &.848 &.825   & .045 &.865 &.847  \\
$Ours$        & \rr{62} (+86)    & \rr{.028} &\rr{.871} &\bb{.837}     &\rr{.025} &\rr{.888} &\rr{.874}      &\rr{.047} &\rr{.882} &\rr{.856}     &\rr{.041}  &\rr{.881} &\rr{.857}   &\rr{.037}  &\rr{.906} &\rr{.889}   \\

\hline
\hline
\end{tabular*}
\end{center}
\vspace{-6mm}
\caption{Quantitative comparisons of with RGB input. The off-the-shelf depth estimation is realized with MiDaS \cite{ranftl2019midas} which presents 86Mb model size. $\uparrow \& \downarrow$ denote larger and smaller is better, respectively. (\rr{red}: best, \bb{blue}: second best).}
\label{tab:rgb}
\vspace{-4mm}
\end{table*}

We firstly compare with RGB models, including R3Net \cite{deng2018r3net}, PoolNet \cite{liu2019simple}, CPD \cite{wu2019cascaded}, AFnet \cite{feng2019attentive}. All saliency maps are directly provided by authors or computed by authorized codes. For fair comparisons, we adopt the same training set as suggested in \cite{piao2020a2dele}, which contains 1485 samples from NJUD, 700 samples from NLPR, and 800 samples from the DUT-RGBD dataset. The remaining images of all listed datasets are used for testing. The quantitative comparison is presented in Table \ref{tab:rgb}. Our model is trained with 50 epochs with $256\times256$ input image size.

For the RGB model, we can conclude from Table \ref{tab:rgb} that the improvement on the saliency map is attributed to different learning modules, which results in high computational cost (size). Different from traditional RGB models which do not exploit the depth information, we propose to take full advantage of the pseudo-geometry estimated with an existing monocular depth estimation method. 

We re-train two RGB-D SOD network (HDFnet \cite{pang2020hierarchical}, CoNet \cite{ji2020accurate}) with the additional estimated pseudo-depth. We observe a significant performance gap between the recent RGB-D models and the previous RGB models. The main reason is the quality of depth estimation: the domain gap between the depth estimation dataset and the SOD dataset leads to some failure depth maps. This can be noticed in the poor performance of HDFnet that extracts features from both RGB and depth images. CoNet, however, is more robust to the depth quality since the depth map is only used to supervise the feature extraction on RGB images. Our model shares the same motivation as CoNet to use depth prior to guide SOD but in a completely different manner. In our model, we directly learn a geometric-aware offset from the depth map to the sampling position on the RGB image. Our model achieves consistent superior performance compared with other models.

\subsection{Performance Comparison with RGB-D Input}

\begin{table*}
\begin{center}
\setlength{\tabcolsep}{1mm}{
\begin{tabular}[ht]{||c l|| c c c | c c c m{1cm}| c c m{1cm} ||}
\hline
\hline
& & \multicolumn{3}{c|}{Extract RGB feature} & \multicolumn{7}{c||}{Extract RGBD feature}\\
\hline
&  & $CoNet_{20}$& $A2dele_{20}$   & $Ours$  & $DANet_{20}$ & $cmMS_{20}$ & $HDFnet_{20}$ & $HDF+Ours$ & $DSA2F_{21}$  &$HDFnet_{20}$ & $HDF+Ours$ \\
\multicolumn{2}{||c||}{Backbone}  & Resnet101 & \multicolumn{2}{c|}{--   VGG16   -- }  & \multicolumn{4}{c|}{--   VGG16   -- }  &\multicolumn{3}{c||}{--   VGG19   --}\\
\hline
& $Size\downarrow$ 
&167 &\underline{57} &  \bl{62}  & \underline{102} & 430 &\underline{177} & \bl{178} & - & \underline{220}  & \bl{221}\\
& $FPS\uparrow$ 
&- &120 &  \bl{150}  & 32 &  &\underline{62} & \bl{58} & - & -  & -\\
\hline
\multirow{4}{*}{\rotatebox[origin=c]{90}{DES}}
& $MAE \downarrow$    
&\underline{.027} &.029 &\bl{.028} &.023 &-  &.030 & \bl{.019} & .021  &.017 & \bl{.017} \\ 
& $F_{\beta}^{mean}\uparrow$ 
&.862 & .870 &\bl{.871} &.887  &-  &.843  &\bl{.920} & .896 &.918  & \bl{.923} \\
& $S_m \uparrow$      
&\underline{.910} & .881 &\bl{.882} &.904 &-  &.899   &\bl{.935}  & .920 &.937 & \bl{.937}\\
& $E_m \uparrow$      
&\underline{.945} & .918 &\bl{.922} &.967 &-  &.944  &\bl{.979}  & .962 &.976 & \bl{.979}\\
\hline
\multirow{4}{*}{\rotatebox[origin=c]{90}{NLPR}}
& $MAE \downarrow$    
&.031 &.031 &\bl{.025} &.028 &.027 &.027 & \bl{.025} & \underline{.024} &.027  & \bl{.025} \\ 
& $F_{\beta}^{mean}\uparrow$ 
&.848 & .871 &\bl{.888} &.871 &.869  &.878 &\bl{.885}  & \underline{.891} &.883 & \bl{.882}  \\
& $S_m \uparrow$      
&.908 & .889 &\bl{.908} &.915 &.899  &.898 & \bl{.918}  & .918 &.915 & \bl{.918} \\
& $E_m \uparrow$      
&.934 & .937 &\bl{.952} &.949 &.945 &.948 &\bl{.954} & .950 &.951 & \bl{.951} \\
\hline
\multirow{4}{*}{\rotatebox[origin=c]{90}{NJUD}}
& $MAE \downarrow$    
&.047 &.052 &\bl{.047} &.045 &.044 &.039 & \bl{.037} & .039  &.038 & \bl{.035} \\ 
& $F_{\beta}^{mean}\uparrow$ 
&.872 & .873 &\bl{.882} &.871 &.886 &.887 &\bl{.893} & .898  &.887 & \bl{.898}  \\
& $S_m \uparrow$      
&\underline{.895} & .867 &\bl{.879} &.899 &.900 &.907  &\bl{.911} & .903 &.911  & \bl{.912}\\
& $E_m \uparrow$      
&.911 & .914 &\bl{.928} &.922 &.914 &.931 &\bl{.935} & .923 &.932 & \bl{.942}\\
\hline
\multirow{4}{*}{\rotatebox[origin=c]{90}{STEREO}}
& $MAE \downarrow$    
&\underline{.037} &.044 &\bl{.041} &.047 &.043 &.042 & \bl{.039}  & .039 &.040& \bl{.039} \\ 
& $F_{\beta}^{mean}\uparrow$ 
&\underline{.885} & .875 &\bl{.881} &.858  &\underline{.879} &.864 &\bl{.864} & \underline{.893}  &.875 & \bl{.878}  \\
& $S_m \uparrow$      
&\underline{.908} & .878 &\bl{.887} &.901 &.895 &.900 &\bl{.904} & .897 &.903  & \bl{.902}\\
& $E_m \uparrow$      
&.928 & .929 &\bl{.936} &.914 &.922  &.929 &\bl{.937} & .933  &.934 & \bl{.938}\\
\hline
\hline
\end{tabular}}
\end{center}
\vspace{-5mm}
\caption{Quantitative comparisons of with recent RGBD models. $\uparrow \& \downarrow$ denote larger and smaller is better, respectively. MGS can also be embedded to the HDFnet \cite{pang2020hierarchical} to enable further progress. The scores/numbers better than ours are \underline{underlined} (extracting RGB feature, extracting RGBD feature with VGG16, and extracting RGBD feature with VGG19 models are labeled separately). More details on all RGB-D dataset can be found in the Supplementary Material. }
\label{tab:rgbd}
\vspace{-1mm}
\end{table*}

We also compare with state-of-the-art RGBD models with raw depth input in the Table \ref{tab:rgbd}, including CoNet \cite{ji2020accurate}, A2dele \cite{piao2020a2dele}, DANet \cite{zhao2020single}, cmMS \cite{li2020rgb}, HDFnet \cite{pang2020hierarchical}, and DSA2F \cite{sun2021deep}. For fair comparisons, all saliency maps and the FPS are directly provided by authors or computed by authorized codes. Note that the FPS depends on the GPU for inference. Thus, only the FPS of HDFnet is tested on the same GPU as ours.

While depth is only used as supervision during training and only RGB image is required during testing, our model surpasses existing efficient A2dele significantly on performance with only an + around 5Mb model size. Compared to CoNet, the model size is minimized by 63\% and achieves a comparable result.  As presented in Figure \ref{fig:guided}, our proposed module can take advantage of cross-modality cues while being lightweight. Thus, we further incorporate with the HDFnet \cite{pang2020hierarchical} to show the performance gain by integrating our approach. It achieves the state-of-the-art (SOTA) performance on VGG16 based models ($HDF+Ours$). To better demonstrate the superiority of the proposed method, we also use a larger backbone (VGG19) to compare with the plain version HDFnet and the SOTA method DSA2F. Note that DSA2F uses neural architecture search to automate the model architecture while ours is hand-designed. Our model enables significant gains on the plain version with minimal cost (+ around 1 Mb on model size) and achieves comparable results with the DSA2F.

\subsection{Qualitative Evaluation}

We present the qualitative result with some challenging cases in Figure \ref{fig:rgb}: low density ($1^{st}$ columns), similar visual appearance between foreground and background ($2^{nd} - 5^{th}$ columns), small objects ($6^{th}$ columns), far objects ($7^{th}-9^{th}$ columns), human in scene ($10^{th}$ columns), and similar and low contrast on depth map ($11^{th}-13^{th}$ columns). It can be seen that our MGSnet yields the results closer to the ground truth mask in various challenging scenarios, especially for the last three columns with low-quality depth clues. Different from two-stream networks that tend to treat sub-optimal depth equally as RGB input, MGSnet extracts features from RGB images while the depth map serves only as complementary guidance, thus becoming robust to depth bias. By analyzing the response on HDFnet (sixth row) and HDFnet with embedded MGS (seventh row), we observe that our approach enables the plain network better discrimination of salient objects from the background.

\section{Ablation Study}

\begin{table*}
\setlength\tabcolsep{0pt}
\setlength\extrarowheight{1pt}
\begin{center}
\begin{tabular*}{0.98\textwidth}{@{\extracolsep{\fill}}*{14}{l}}
\hline
& Dataset   &\multicolumn{2}{c}{DES} &\multicolumn{2}{c}{LFSD} & \multicolumn{2}{c}{NLPR}  & \multicolumn{2}{c}{NJUD}  &\multicolumn{2}{c}{SSD}  & \multicolumn{2}{c}{STEREO} \\
\cline{3-4} \cline{5-6} \cline{7-8} \cline{9-10} \cline{11-12} \cline{13-14}
\# & Metric &$MAE\downarrow$ &  $F_{\max}\uparrow $  &$MAE\downarrow$ &  $F_{\max}\uparrow $  &$MAE\downarrow$ &  $F_{\max}\uparrow $  &$MAE\downarrow$ &  $F_{\max}\uparrow $  &$MAE\downarrow$ &  $F_{\max}\uparrow $  &$MAE\downarrow$ &  $F_{\max}\uparrow $ 
\\
\hline

1 & RGB Baseline
& .036 & .880    &.104 &.835   &.037  &.883  &.059  &.877    &.082  &.785    &.054 &.882  
\\
2 & RGB + Self Deform
& .042 & .860    &.110 &.798   &.031  &.885  &.052  &.879    &\underline{.062}  &\underline{.814}    &.046 &.882  
\\
3 & RGB pseudo D Fusion
& .032 & .888    &.093 &.819   &.029  &.893  &.061  &.863    &.077  &.776    &.049 &.878  
\\
4 & RGB + Depth-Deform
& \bl{.028} & \bl{.899}    &\bl{.078} &\bl{.849}   &\bl{.025}  &\bl{.905}  &\bl{.047}  &\bl{.897}    &\bl{.063}  &\bl{.801} &\bl{.041} &\bl{.898}  
\\

\hline
5 & RGB-D Baseline (B)
& .020 & .933    &.089 &.856   &.028  &.921  &.039  &.922    &.054  &.867 &.042 &.911  
\\
6 & B + Self Deform
& .019 & .936    &.089 &.853   &.026  &.916  &.038  &.925    &.053  &.878    &.044 &.895  
\\
7 & B + Cross-Modal Deform
& \bl{.019} & \bl{.936}    &\bl{.079} &\bl{.871}   &\bl{.025}  &\bl{.921}  &\bl{.037}  &\bl{.925}    &\bl{.049}  &\bl{.867} &\bl{.039} &\bl{.917}  
\\
\hline
\hline
\end{tabular*}
\end{center}
\vspace{-6mm}
\caption{Ablation study of modality-guided sampling position}
\label{tab:ablation}
\vspace{-4mm}
\end{table*}

\textbf{Effect of Modality-Guided Sampling Position}: Our modality-guided sampling position aims to incorporate multi-modal information through the basic function of CNN - the sampling position of convolution. This pattern is integrated in Eq. \ref{eq} and Eq. \ref{eq:offset}. To verify the effectiveness of the proposed modality-guided sampling position, a series of experiments with different learning strategies are realized. 

(1) - (4) are experiments on RGB model:  (1) RGB Baseline. (2) Self-guided deformable sampling position. We learn the offset from the RGB feature map. (3) RGB pseudo-depth early fusion. We form a four-channel input with pseudo depth. (4) Depth-guided deformable position. We compute an offset from pseudo-depth using Z-ACN to guide RGB streaming.  (5) - (7) are experiments on RGBD model: (5) Baseline. We use the same architecture as HDFnet. (6) Self-guided deformable sampling position. The offset applied to RGB streaming is learned from the RGB feature. Idem for depth streaming. (7) Cross modality-guided deformable position. We learn an offset from depth to guide RGB streaming, and vice versa. 

\begin{figure}[ht]
\centering
\vspace{-1mm}
\includegraphics[scale = 0.61]{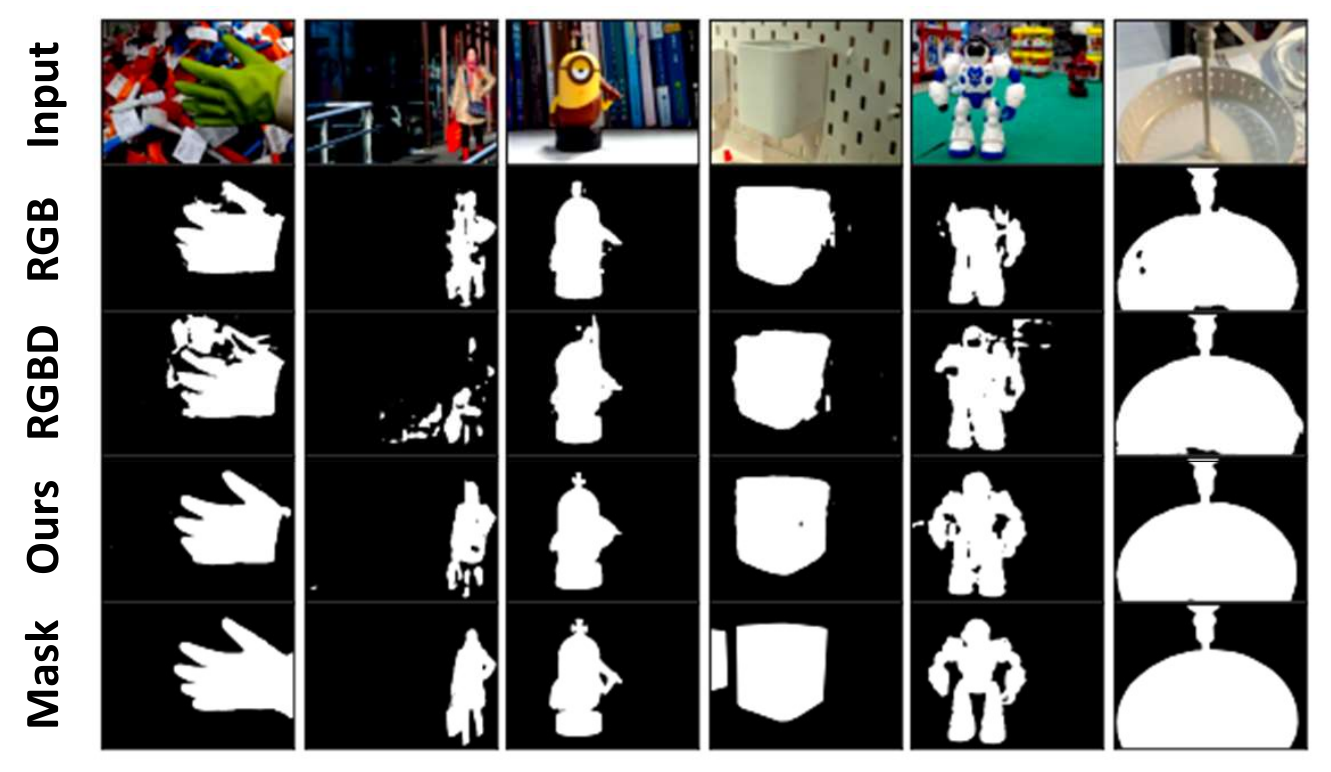}
\vspace{-4mm}
\caption{Visual analysis of embedded depth with MGSnet.}
\label{fig:data}
\vspace{-5mm}
\end{figure}

Table \ref{tab:ablation} (1) and (3) compare the performance of the baseline RGB three-channel input and mimicked RGBD four-channel input with pseudo-depth, respectively. The mimicked multi-modality early fusion achieves better performance, indicating that the pseudo-depth provides additional semantic. However, by comparing (3) and (4), we observe that the proposed depth-guided deformable sampling position can better use the complementary information to supervise RGB streaming, compared with early fusion. By comparing (2) and (4), we show that the depth-guided deformable position is more accurate on saliency compared to that of the self/RGB-guided. This verifies the assumption that depth cues can help the RGB model to better distinguish the foreground and background. Note that in (4) we only extract features from RGB images. The additional awareness of the geometry is only treated as a 2D offset to better locate the sampling position. This new integration design contributes to the model performance with minimal cost. For better understanding, the qualitative result presented in Figure \ref{fig:data} shows that our approach provides more accurate saliency maps with better contrast. On the RGBD model (5-7), we also observe the superior performance with the cross-modality deformable sampling position achieves as it directly compensates for the single modal streaming.

\textbf{Performance with different depth qualities}:
We also conduct an experiment to show the impact of depth quality. We choose the HDFnet \cite{pang2020hierarchical} as the baseline and further embed it with our method. We present the average metric on all testing datasets in Table \ref{tab:estimated} with pseudo-depth (estimated) and raw depth from the RGBD dataset. Results obtained with pseudo-depth are denoted with *. 

\begin{table}[ht]
\begin{center}
\begin{tabular}[ht]{l | c c | c  c }
\hline
\small{AvgMetric} & $HDFnet*$ & $+Ours*$ & $HDFnet$ & $+Ours$ \\
\hline
 $MAE \downarrow$    
&.1053 &\rr{.0758}  & .0405 & \rr{.0375}  \\ 
 $F_{\beta}\uparrow$ 
&.8410 &\rr{.8599} & .9121 & \rr{.9166}\\
 $F_{\beta}^{mean}\uparrow$ 
&.7326 &\rr{.7868} & .8730 & \rr{.8831} \\
 $F_{\beta}^{w}\uparrow$ 
&.6789 &\rr{.7488}  & .8569 & \rr{.8672} \\
 $S_m \uparrow$      
&.8010 &\rr{.8390} & .9013 & \rr{.9053} \\
 $E_m \uparrow$      
&.8359 &\rr{.8797} & .9312 & \rr{.9377} \\
\hline
\hline
\end{tabular}
\end{center}
\vspace{-4mm}
\caption{Performance variation with different depth qualities. (*) denotes results obtained with pseudo-depth.}
\label{tab:estimated}
\vspace{-5mm}
\end{table}

It shows that the quality of depth has an important influence on performance. Features extracted from raw depth describe better the salient object and were in line with our expectations. However, in both cases, our MGS can significantly enable progress compared to the plain networks. For pseudo-depth, the contribution of our MGS is more significant, which can be explained by the effectiveness of our RGB-guided sampling position for depth streaming. It can efficiently help to alleviate depth errors.

\section{Conclusions}

In this paper, we propose a modality-guided module (MGSnet) for RGB-D salient object detection models. The depth channel can either be the input or be estimated using a state-of-the-art monodepth network. Our model adaptively transforms convolutions that their size and shape are built by fusing information from one modality to another (e.g., depth to RGB and/or RGB to depth), thus enabling a cross-modality interaction. Extensive experiments against RGB baselines demonstrate the performance gains of the proposed module, and the addition of the proposed module to existing RGB-D models further improved results.

\section*{Acknowledgements}

We gratefully acknowledge Zhuyun Zhou for her support and proofreading. We also thank Jilai Zheng, Zhongpai Gao, and Yonglin Zhang for the discussion. This research is supported by the French National Research Agency through ANR CLARA (ANR-18-CE33-0004) and financed by the French Conseil R\'egional de Bourgogne-Franche-Comt\'e. 

{\small
\bibliographystyle{ieee_fullname}
\bibliography{egbib}
}

\end{document}